%% file: draft.tex
\documentclass[letterpaper, 10 pt, conference]{ieeeconf}  % Comment this line out if you need a4paper

\IEEEoverridecommandlockouts                              % This command is only needed if 
\usepackage{amsmath,amsfonts}

\let\labelindent\relax
\usepackage{amssymb, amsthm}
\usepackage{mathtools}
\usepackage{dblfloatfix}

\usepackage{calc}
\usepackage{cases}
\usepackage{url}
\usepackage{hyperref}

\usepackage{float,color,graphicx}
\usepackage{tikz}
\usepackage{verbatim}
\usetikzlibrary{calc}
\usetikzlibrary{patterns}
\usetikzlibrary {arrows.meta}
\usepackage{pgfplots}
\pgfplotsset{compat=1.15}

\usepackage{enumitem}
\usepackage[font=footnotesize,skip=4pt]{caption}
\usepackage{cite}
\usepackage{balance}
\usepackage{tabularx}
\usepackage[position=t,singlelinecheck=off,caption=false]{subfig}
\usepackage{multirow}
\usepackage{enumerate}

\usepackage{mymath}
\usepackage{SLAM}
\usepackage{slam_methods}

\newcommand{\ignore}[1]{}

\title{\LARGE \bf
Good Weights: Proactive, Adaptive Dead Reckoning Fusion \\ for Continuous and Robust Visual SLAM 
}

\author{{Yanwei Du, Jing-Chen Peng,  Patricio A. Vela}
\thanks{ {*Supported in part by NSF Awards \#2235944, \#2345057.} }% <-this % stops a space
\thanks{Y.~Du, J.~Peng and P.~A.~Vela are with the 
School of Electrical and Computer Engineering, and 
Institute of Robotics and Intelligent Machines,  
Georgia Institute of Technology, Atlanta, GA 30308, USA.
{\tt\small \{yanwei.du, jpeng303, pvela\}@gatech.edu}}
}
\begin{document}

\maketitle
\thispagestyle{empty}
\pagestyle{empty}

%%%%%%%%%%%%%%%%%%%%%%%%%%%%%%%%%%%%%%%%%%%%%%%%%%%%%%%%%%%%%%%%%%%%%%%%%%%%%%%%
\begin{abstract}
Given that Visual SLAM relies on appearance cues for localization and scene
understanding, texture-less or visually degraded  environments (\textit{e.g.},
plain walls or low lighting) lead to poor pose estimation and track loss.
However, robots are typically
equipped with sensors that provide some form of dead reckoning odometry with 
reasonable short-time performance but unreliable long-time performance.
The \textit{Good Weights} (\gw) algorithm described here provides a framework
to adaptively integrate dead reckoning (DR) with passive visual SLAM for
continuous and accurate frame-level pose estimation.  Importantly, it describes
how all modules in a comprehensive SLAM system must be modified to incorporate
DR into its design. 
Adaptive weighting increases DR influence when visual tracking is unreliable and
reduces when visual feature information is strong, maintaining pose track without
overreliance on DR.  \textit{Good Weights} yields a practical solution for
mobile navigation that improves visual SLAM performance and robustness. 
Experiments on collected datasets and in real-world deployment demonstrate
the benefits of \textit{Good Weights}.
% \textcolor{red}{
% Experiments on public and self-collected datasets show that our method 
% matches active depth-based SLAM and outperforms state-of-the-art DR-aided 
% stereo visual SLAM approaches.
% }

% TODO (yanwei) Releasing code before reviewing is against double-blind???
% \textcolor{red}{
% The implementation is open-source to support the community.
% }

\textbf{Keywords:} Visual SLAM, dead reckoning, feature tracking, optimization

\end{abstract}

%%%%%%%%%%%%%%%%%%%%%%%%%%%%%%%%%%%%%%%%%%%%%%%%%%%%%%%%%%%%%%%%%%%%%%%%%%%%%%%%
\section{Introduction \label{sec:intro}}
\input{intro.tex}

%%%%%%%%%%%%%%%%%%%%%%%%%%%%%%%%%%%%%%%%%%%%%%%%%%%%%%%%%%%%%%%%%%%%%%%%%%%%%%%%
\section{Related Works \label{sec:back}}
\input{lr.tex}

%%%%%%%%%%%%%%%%%%%%%%%%%%%%%%%%%%%%%%%%%%%%%%%%%%%%%%%%%%%%%%%%%%%%%%%%%%%%%%%%
\section{Methodology \label{sec:meth}}
\input{methodology.tex}

%%%%%%%%%%%%%%%%%%%%%%%%%%%%%%%%%%%%%%%%%%%%%%%%%%%%%%%%%%%%%%%%%%%%%%%%%%%%%%%%
\section{Experiments \label{sec:exp}}
\input{exp.tex}

%%%%%%%%%%%%%%%%%%%%%%%%%%%%%%%%%%%%%%%%%%%%%%%%%%%%%%%%%%%%%%%%%%%%%%%%%%%%%%%%
\section{Conclusion \label{sec:conclusion}}
\input{conclusion.tex}

% \newpage
\bibliographystyle{IEEEtran}
\balance
\bibliography{bibliography}

\end{document}

%% file: intro.tex
%% General Context
% Visual SLAM is a core technology for mobile robots, supporting everyday tasks such as navigation, delivery, and interaction in indoor environments \cite{bujanca2021robust, Barakeh2019PepperHR}.
% % Cameras are lightweight and energy efficient, with a small form factor that makes them easy to mount on mobile robots without complex calibration. \cite{TODO}
% In real world deployment, robots often face textureless scenarios or poor lighting, where reliable features are scarce. In these conditions, visual tracking becomes fragile, leading to drift or even complete failure with no available visual measurements \cite{wang2025recent}.

Visual Simultaneous Localization and Mapping (SLAM) is often formulated as a nonlinear least-squares problem, where camera poses and 3D landmarks are jointly estimated from visual observations\cite{murORB2, campos2021orb, zhao2020good}. 
Optimization accuracy and stability depends on the sufficiency and reliability of feature associations across frames, 
short-term and long-term.
% When associations are accurate and priors are consistent, bundle adjustment and pose graph optimization yield highly precise reconstructions [Agarwal et al., 2010]. 
%When associations fail in low-texture or ambiguous environments, or when 
Incorrect motion priors, low-texture scenes, and other visually ambiguous situations result in poor or missing data association for which the
least-squares problem diverges or fails \cite{wang2025recent, bujanca2021robust}. 

%% What are the most common approaches and limitations.
Dead reckoning, typically from IMU and wheel encoders \cite{wu2022ral}, supplements visual odometry with continuous pose estimates \cite{qin2017vins, mourikis2007multi, Geneva2020OpenVINSAR, Rosinol19arxiv-Kimera, Zheng2025Wheel}.
These measurements are integrated either through pre-integration or direct pose factors in tightly- or loosely-coupled formulations.
Running at higher frequency than cameras, they capture robot kinematics and supply short-term motion priors that help maintain localization when visual features are weak or unavailable. 
However, such systems fuse measurements with fixed weights, making them sensitive to special motion patterns and often dependent on online calibration to compensate for imperfect modeling \cite{Wu2017VINSOW, Lee2020VisualInertialWheelOW}. 
More importantly, these methods remain odometry-focused rather than integrated into a full visual SLAM pipeline, where map construction can reinforce visual constraints rather than relying predominantly on inertial cues.

\begin{figure}[t]
  \vspace{0.06in}
  \centering
  \begin{tikzpicture}[inner sep=0pt, outer sep=0pt]
    \node[anchor=north west] (cid_apart_traj) at (0,0) 
      {\includegraphics[width=0.34\textwidth]{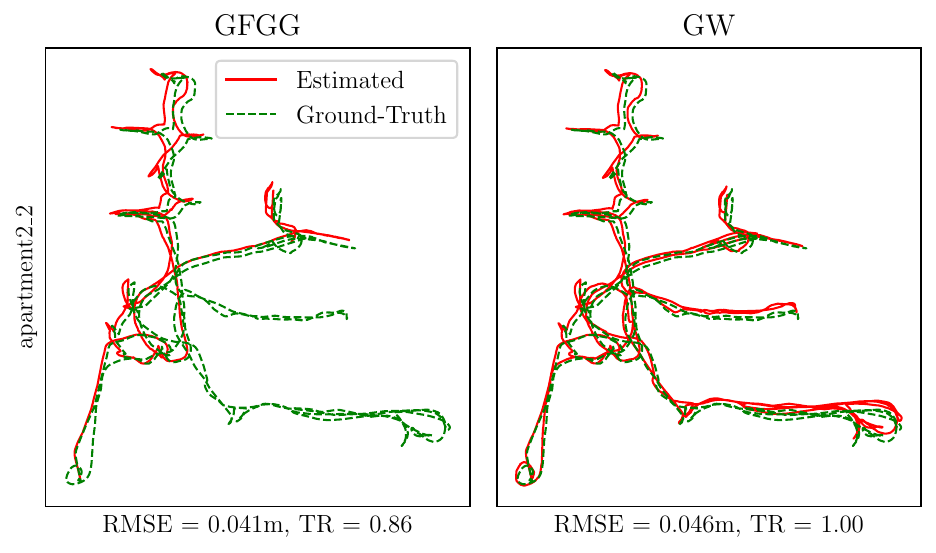}};
    \node[anchor=east] (cid_apart_im) at ($(cid_apart_traj.west)+(0em, 1.0em)$)
      {\includegraphics[width=0.15\textwidth, trim={0 3cm 0 0}, clip]{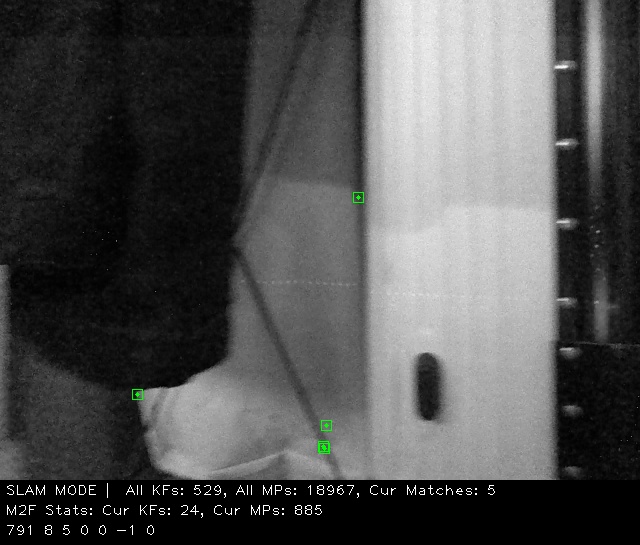}};
    \node[anchor=north] (cid_floor_im) at ($(cid_apart_im.south)+(0em, -1.0em)$)
      {\includegraphics[width=0.15\textwidth, trim={0 3cm 0 0}, clip]{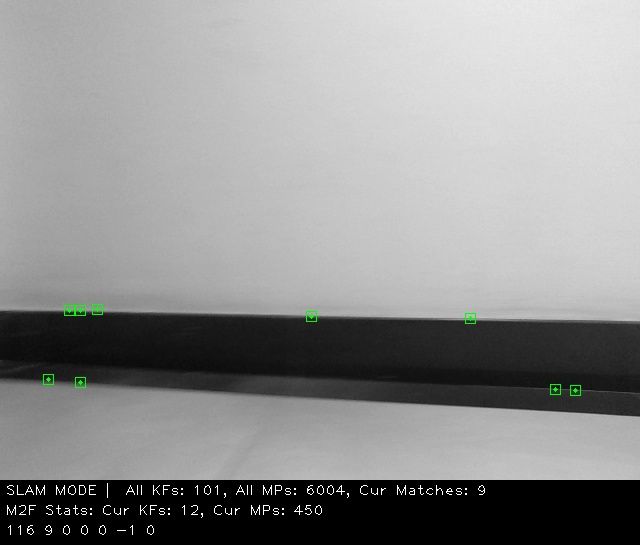}};
    \node[anchor=north] (cid_floor_traj) at ($(cid_apart_traj.south)+(0.0em, 0.0em)$)
      {\includegraphics[width=0.34\textwidth]{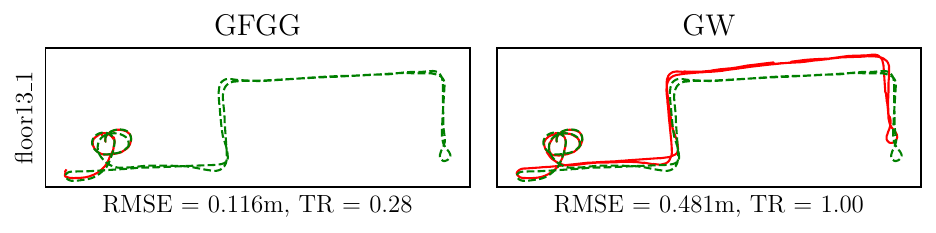}};

    % Left side captions
    \node at ($(cid_apart_im.south)+(0em, -0.5em)$) {\tiny Tracked Points: 5};
    \node at ($(cid_floor_im.south)+(0em, -0.5em)$) {\tiny Tracked Points: 9};

    % Callouts to the points in the trajectory
    % Define target points.
    % Adjust these to adjust the callout positions.
    \coordinate (t1) at ($(cid_apart_traj.center)+(-0.8cm, 0.25cm)$);
    \coordinate (t2) at ($(cid_floor_traj.center)+(-2.5cm, -0.35cm)$);

    % Color for all markers
    \newcommand{\colorA}{blue}
    
    % Lines from top
    \coordinate (s1) at ($(cid_apart_im.north east)$);
    \coordinate (s2) at ($(cid_apart_im.south east)$);
    \coordinate (c) at ($(t1)$);
    \draw[color=\colorA, -] ($(s1)$) -- ($(c)$);
    \draw[color=\colorA, -] ($(s2)$) -- ($(c)$);
    \draw[color=\colorA, fill=white] ($(c)$) circle (0.05);
    
    % Lines from bottom
    \coordinate (s1) at ($(cid_floor_im.north east)$);
    \coordinate (s2) at ($(cid_floor_im.south east)$);
    \coordinate (c) at ($(t2)$);
    \draw[color=\colorA, -] ($(s1)$) -- ($(c)$);
    \draw[color=\colorA, -] ($(s2)$) -- ($(c)$);
    \draw[color=\colorA, fill=white] ($(c)$) circle (0.05);
    
    % \draw[red, dashed, ->] (1,0) -- (1,3);
  \end{tikzpicture}
  \caption{
  Performance comparison between a visual SLAM method, {\gfgg}, and its \textit{Good Weights} augmentation, GW, on the CID dataset for the apartment (top) and floor (bottom) sequences. The left-most images show instances of poor visual signals with an indication of their location in the trajectory (blue circle).
  The GW augmentation supports full tracking and SLAM success in these scenarios where the standard approach fails.\label{fig:cid_traj}}
  \vspace{-15pt}
\end{figure}
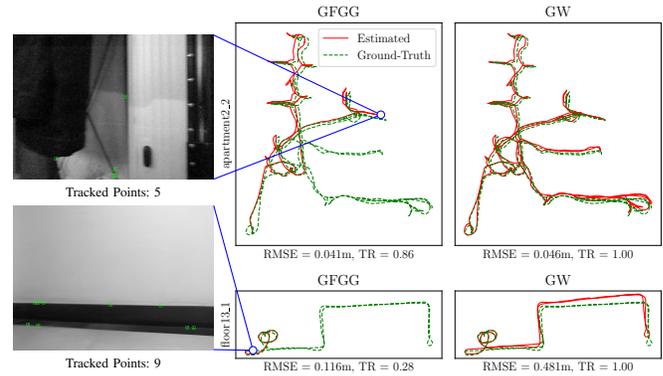

%% What are most recent approaches and limitations.
Recent studies \cite{Pritzl2023Adaptive, Vsevolod2024Adaptive, yin2024ground, Jia2021LvioFusionAS, Jaafar2023Wheel} explore dynamic tuning of constraint weights to prevent one sensor from dominating the optimization. They highlight the benefit of treating sensor contributions as variable rather than fixed influences.
% \textcolor{red}{
% The weighting for visual measurements usually on XXX.
% }
An extreme form of adaptive weighting leads to a switching SLAM \cite{Lee2024SwitchSLAMSL, Xu2024SelectiveKF}, where the system selects one modality as the main source and switches to others during sensor degradation.
Most switching designs focus on LiDAR-based SLAM, where degenerate environmental structure (e.g., corridors, tunnels) is detected via scan-matching residuals or Jacobian conditioning \cite{Tuna2024InformedCA, Zhang2016OnDO, Zhao2024SuperLocTK, Han2023DAMSLIOAD}. 
% Extending these ideas to visual SLAM is still underexplored. 
LiDAR-based pipelines, with direct geometric measurements, can apply post-optimization corrections such as reweighting or smoothing.
In contrast, indirect visual SLAM maintains a multi-stage hierarchy that relies on appearance-based cues, not structural cues.
Its design inherently involves multiple dependent modules: tracking, local mapping, and loop closing, each of which requires the previous stage to remain well-conditioned. 
When visual signals are degraded, failures in early tracking or local mapping propagate through the pipeline and compromise global consistency.

% Observability depends on feature detection and tracking at the front end, followed by pose estimation and optimization at the back end.
% When associations are weak, the jacobians become ill-conditioned or invalid, and post-hoc reweighting or damping cannot recover information that was never present. 
% Moreover, these visual factors are difficult to model accurately, and reliable tracking metrics remain an open challenge.

The impact of poor visual signals motivates the use of dead-reckoning (DR) motion priors as a structural component in visual SLAM for robots. 
DR provides pre-conditioning across modules to handle degenerate scenarios that cannot be addressed by post-hoc reweighting.
% This motivates the need for strategies that balance auxiliary motion priors with visual tracking in challenging indoor environments. 
This paper introduces a proactive, adaptive DR-aided visual SLAM framework. 
When feature association becomes poor or fails, the system incorporates DR as a motion prior to maintain pose continuity.
Unlike tightly coupled fusion methods, DR is only used when needed and the system smoothly returns to vision once tracking recovers.
This design uses lightweight visual-health indicators to regulate DR contributions and preserve long-term accuracy.
The framework, denoted \textit{Good Weights}, provides the following contributions:
\begin{itemize}[leftmargin=*]
    \item A proactive, adaptive weighting scheme using tracking quality to modulate dead reckoning as a motion prior, 
      leading to improved robustness under poor visual tracking.
    % \item A proactive, adaptive regulation strategy that adjusts dead reckoning weighting based on real-time tracking observability, enabling stable motion prior integration during visual degradation.
    \item A comprehensive strategy integrating dead reckoning priors into tracking, local mapping, and loop closing, while maintaining 
      real-time operation.
    %\item Experiments on indoor datasets with low-texture conditions and on the \textcolor{red}{real robot} demonstrate our approach improves continuity and stability without sacrificing accuracy under normal conditions.
    \item Experiments on indoor datasets and on a mobile robot demonstrate improved continuity and stability in low-texture conditions without sacrificing accuracy under normal conditions.
\end{itemize}

% In our work, we propose an adaptive dead reckoning fusion framework with passive visual SLAM.

% The system is built around a passive visual SLAM pipeline, ensuring high accuracy and long-term consistency under normal conditions.
% DR is introduced adaptively as a motion prior when visual tracking becomes poor or fails completely (e.g., when feature association cannot provide reliable constraints).
% Unlike tightly coupled fusion frameworks, our method emphasizes conditional reliance: DR contributes only when needed, preventing over-weighting while ensuring continuity during visual failures.
% Once visual tracking recovers, the system smoothly returns to using visual SLAM as the main source of pose estimation, keeping long-term drift low while maintaining robustness.
% The approach avoids complex wheel or inertial modeling, focusing instead on adaptive weighting based on visual-health indicators (feature counts, inlier ratios, residuals). This makes the system simple, efficient, and portable across mobile robot platforms.

% What are the gaps and what are the contributions of the work.

%% file: lr.tex
\subsection{Dead Reckoning-aided Visual SLAM/Odometry}

Inertial and wheel measurements show promising performance improvements when fused with visual observations \cite{campos2021orb, mourikis2007multi, Geneva2020OpenVINSAR, chen2023stereo}, typically within factor-graph or filtering frameworks. Fusion covariance parameters are derived from sensor specifications, feature detection and tracking statistics, or learned uncertainty models. 
These fusion techniques emphasize short-horizon odometry estimation and do not fully exploit the visual map constructed during robot navigation.
Yet this map provides valuable visual constraints through landmark re-observations and loop closures, which improve accuracy and optimization conditioning \cite{mur2017visual, du2024task}.
\textit{Given the benefit of longer-term visual association,  weighting strategies are needed 
%consider motion priors at the trajectory level and 
for the stronger structural constraints provided by the visual map.}

\subsection{Adaptive-Fusion SLAM}
Adaptive fusion approaches for improved pose estimation in challenging scenarios broadly divide into two categories.
The first follows the classical multi-sensor design \cite{lynen2013robust, Nubert2025HolisticFT}, 
in which all available modalities are fused and their weights are adjusted adaptively through heuristic rules or learning-based schemes\cite{Jia2021LvioFusionAS}. 
% Reliability is typically inferred from abnormality detection or information-theoretic measures.
% with weights tuned either through heuristic rules or learning-based schemes \cite{chuang2023into, Jia2021LvioFusionAS}.
% These methods reduce manual tuning effort and improve resilience to corner cases. However, they focus mainly on down-weighting failed sensors and often overlook the reverse case: when one modality provides strong and reliable measurements, it should dominate the estimation instead of being averaged with others.
The second category adopts switching-based fusion, where one primary modality drives estimation and backup modules are activated when degenerate conditions are detected \cite{Lee2024SwitchSLAMSL, Xu2024SelectiveKF}. 
The strategy is effective in LiDAR SLAM, where degeneracy is identified during pose optimization by analyzing Jacobian conditioning \cite{Zhang2016OnDO}. 
Extending this idea to visual SLAM is non-trivial: due to its hierarchical
design, conditioning analysis depends on reliable tracking and feature associations, 
which may already have failed prior to optimization in low-texture scenarios.
Other LiDAR works exploit points measurements with motion profiles at earlier stages \cite{Zhao2024SuperLocTK, Junior2022EKFLOAM}. 
These approaches rely on active depth sensing, which measures geometry directly. 
In contrast, (passive) visual SLAM must indirectly infer geometry from appearance cues, 
which are less consistently correlated with the underlying structure, making the problem inherently more difficult.
% \textcolor{blue}{%
% In contrast, passive visual measurements do not provide strong system conditioning\cite{Demmel2021SquareRM}, 
% making the approach fundamentally more difficult in visual SLAM.}

% Analytical conditioning in visual SLAM for reliability estimation remains difficult. As the feature association is fragile and unstable, and even absent in texture-less scenes, meaning jacobians alone do not reflect the true system performance\cite{Demmel2021SquareRM}. 
% Besides, a dropping in tracked features does not always indicate failures, in theory, as few as three points can be sufficient for pose estimation (P3P).

\subsection{Optimization}
Adaptive strategies exist in optimization \cite{Nocedal2018NumericalO}:
Levenberg–Marquardt (LM) \cite{ranganathan2004levenberg} and trust-region approaches \cite{conn2000trust} dynamically adjust a damping factor to trade off between gradient-descent-like steps and Gauss–Newton updates. 
Varying optimization weights based on the outcome of the previous optimization step leads to improved convergence. 
Similar ideas appear in robust estimation, where residuals are down-weighted online using kernels such as Huber or Cauchy \cite{triggs1999bundle}.
\textit{Good Weights} follows a similar design but anticipates unreliable measurements before optimization, scaling the contribution of dead reckoning according to visual tracking quality indicators. 
\textit{It complements post-hoc adaptive schemes by addressing failure cases earlier in the pipeline, ensuring that the underlying Hessian remains well conditioned even when visual constraints are scarce.}

\vspace*{1\baselineskip}
In summary, prior methods in optimization (LM, trust region) and sensor fusion (DR-aiding with adaptive weighting) share a reactive nature: 
they stabilize optimization only after evaluating progress.
LiDAR systems can afford this because dense measurements ensure that residuals remain informative.
Visual SLAM, however, cannot. 
\textit{Good Weights} addresses this gap by differentially employing DR priors across the SLAM hierarchy to
maintain well-conditioned subproblems throughout the pipeline.
This proactive, hierarchical strategy makes \textit{Good Weights} distinct from multi-sensor smoothing and classical damping techniques.

% Our work extends these ideas to Visual SLAM, with a focus on handling visual tracking loss or failure. We develop an adaptive scheme that adjusts the influence of dead reckoning based on tracking quality, enabling smooth transitions between DR and vision. This design maintains stable estimation during failures while preserving accuracy when visual cues are reliable.

% These limitations suggest that reliability in visual SLAM cannot be captured by fixed analytical metrics alone. Instead, robustness requires strategies that adapt sensor influence based on the actual tracking condition. Dead reckoning provides a natural complement, offering stable short-term motion priors when vision becomes unreliable.

% TODO (yanwei): already mentioned at the end of introduction ???
% Building on this insight, we propose an adaptive weighting framework that increases the role of dead reckoning during visual tracking failures and suppresses it when vision is stable, ensuring both continuity and accuracy in pose estimation.

%% file: methodology.tex
\subsection{System Overview}

\begin{figure}[t]
  % \vspace{0.06in}
  \centering
  \includegraphics[width=1.0\linewidth]{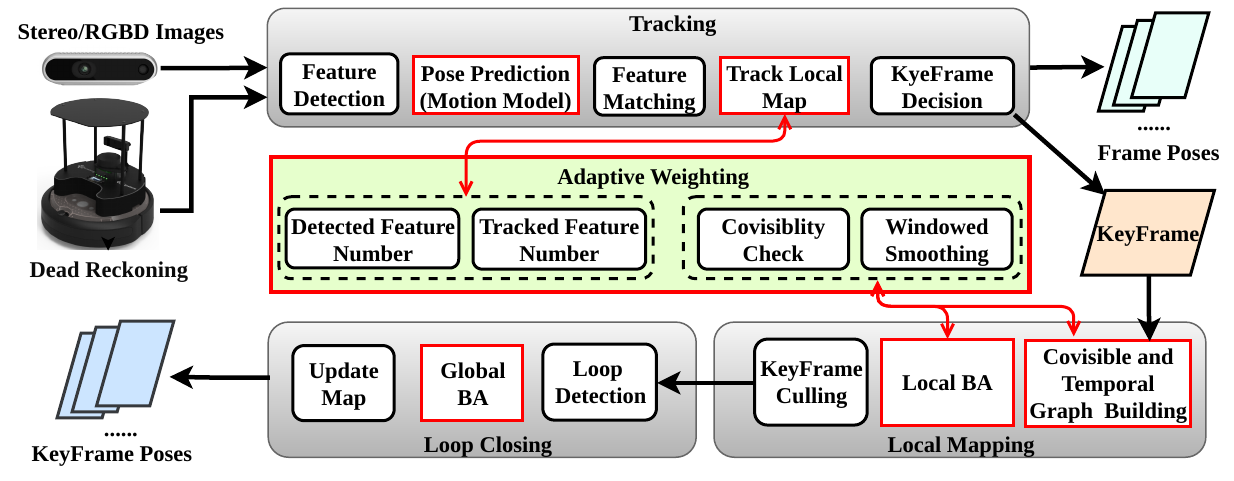}
  \caption{System overview following the design of \cite{murORB2, zhao2020good}, with newly added and modified modules highlighted in red rectangle blocks. The main contribution is the adaptive weighting scheme, where dead reckoning (DR) is selectively introduced as a motion prior within the system.\label{fig:sys_overview}}
  \vspace{-15pt}
\end{figure}

The framework builds on visual SLAM pipeline \cite{zhao2020good}, which is organized into three main threads: tracking, local mapping, and loop closing, as illustrated in Fig.~\ref{fig:sys_overview}.
Frame poses are estimated in the tracking thread at the image-stream rate, providing real-time localization for online operation.
Keyframes are initialized from selected frames, and their poses are further optimized in a local map or a global map depending on whether loop closure is triggered, reflecting post-processed mapping accuracy.
Vision remains the primary source of pose estimation, ensuring accuracy and long-term consistency when features are reliably tracked.
To handle situations where feature association becomes weak or fails, the system incorporates dead reckoning (DR) as an auxiliary motion prior.
The primary contribution of this work is an adaptive weighting scheme that regulates how much influence DR has at each stage of the pipeline.
When visual tracking is strong, DR plays a minor role; when tracking degrades, DR stabilizes the estimate; and when vision fails entirely, DR maintains continuity until recovery.
The following subsections first introduce the adaptive weighting formulation, then describe its integration into the tracking, local mapping, and loop closing modules.

% The pipeline supports both stereo and RGB-D image streams. In RGB-D mode, depth is only used to initialize landmarks when a new keyframe is created. This design contrasts with other RGB-D SLAM/Odometry systems that involve 3D–3D registration \cite{rgbd-slam}.
% The DR input is obtained from the onboard robot, typically robot odometry or IMU-based odometry, and can be integrated with any standard fusion framework as \cite{ekf}. 
% The central idea of our adaptive fusion is straightforward: 
% when visual tracking is reliable, the system relies primarily on visual SLAM; 
% when tracking degrades, DR contributes more strongly; 
% and when vision fails completely, DR alone propagates the pose until visual tracking recovers.
% The following sections introduce our method and the role of motion priors within each module.

\subsection{Adaptive Weighting Scheme}
In what follows, all frame poses, including motion priors from DR, are expressed in the camera frame for consistency, assuming that sensor calibration and time synchronization have been properly handled.
\subsubsection{Problem Formulation}
The goal of visual SLAM is to estimate the sequence of frame poses $\mathbf{T}_{1:T} \in \mathrm{SE}(3)$ given image measurements and auxiliary motion priors.
Without loss of generality, express the SLAM bundle adjustment form as:
\begin{equation}
\min_{\mathbf{T}_{1:T}, \mathbf{X}} \;
\underbrace{\sum_{(i,j)} 
\left\| r^{\text{vis}}_{ij}(\mathbf{T}_i, \mathbf{X}_j) \right\|^2_{\Sigma_v^{-1}}}_{\text{visual reprojection residuals}}
+ 
\underbrace{\sum_{k} 
\left\| r^{\text{DR}}_{k}(\mathbf{T}_{k}, \mathbf{T}_{k-1}) \right\|^2_{\Sigma_d^{-1}}}_{\text{dead-reckoning priors}},
\label{eq:ba}
\end{equation}
where $r^{\text{vis}}_{ij}$ denotes the reprojection error of landmark $j$ in frame $i$, 
and $r^{\text{DR}}_{k}$ denotes the relative pose residual 
from DR between consecutive frames $k-1$ and $k$.
The covariances $\Sigma_v$ and $\Sigma_d$ determine the weighting of each contribution.

% In the tracking thread, we employ motion-only BA for pose estimation, which can be viewed as a reduced form of the same optimization problem as in Eq.~\eqref{eq:ba}. Here, only the current camera pose is optimized while landmark positions are held fixed:
% \begin{equation}
% \min_{\mathbf{T}_t}
% \sum_{j} \left\| r^{\text{vis}}_{tj}(\mathbf{T}_t, \mathbf{X}_j) \right\|^2_{\Sigma_v^{-1}}
% + 
% \left\| r^{\text{DR}}_{t}(\mathbf{T}_{t}, \mathbf{T}_{t-1}) \right\|^2_{\Sigma_d^{-1}}
% \label{eq:mba}
% \end{equation}

% From the linearized system of Eq.~\eqref{eq:ba}, the Hessian takes the form (ignoring the subindex)
% \begin{equation}
%     H = J_v^T \Sigma_v^{-1} J_v + J_d^T \Sigma_d^{-1} J_d \label{eq:hess}
% \end{equation}

In practice, visual information is not constant. When enough good features are tracked, visual residuals are reliable and dominate the optimization. When features are few or mismatched, the visual Jacobian ${J_v}$ becomes weak, yielding an ill-posed optimization problem \cite{triggs1999bundle}.
DR provides continuous motion estimates but suffers from drift over long horizons. If DR is always weighted the same, it either dominates and pulls the estimate off track, or becomes too weak to help when vision fails.

This motivates the use of an adaptive scheme. Rather than altering the visual covariances which would require complex per-feature uncertainty modeling \cite{Qiu2024MACVOMC} and extensive tuning as often seen in sensor fusion approaches \cite{ebadi2023present, Nubert2025HolisticFT}, 
% we keep $\Sigma_v$ fixed and instead regulate the contribution of DR through a global measure of visual tracking quality.
the visual covariance $\Sigma_v$ is kept fixed, and the contribution of DR is regulated through a global measure of visual tracking quality.

% TODO (yanwei): evidence of why the covariance needs tuning??? 1) imperfect modeling, non-gaussian noise. 2) sensor degradation.
% The challenge is to decide when and how strongly to rely on DR: $\Sigma_v^{-1}$ v.s. $\Sigma_d^{-1}$. If we give DR too much influence, it will dominate and introduce drift; if we trust it too little, it cannot prevent failure when vision drops out.

% Proper modeling of uncertainty, and the ratio of that matters.

\subsubsection{Adaptive Weighting Formulation}
The score $Q_t \in [ 0, 1]$ is introduced to represent the health of visual tracking at time $t$.
A value of $Q_t = 1$ indicates strong and stable tracking, while $Q_t = 0$ means no visual constraints are available.
The DR prior information is scaled as
\begin{equation} \label{eq:dr_cov}
\Sigma_d^{-1}(t) = \alpha(Q_t) \, \Sigma_{d0}^{-1},
\end{equation}
where $\Sigma_{d0}^{-1}$ is the nominal DR information matrix. 
The DR weight is interpolated $\alpha$ between lower and upper bounds
$\alpha_{\min}$ and $\alpha_{\max}$ as a function of the tracking quality $Q_t$.
% \in [0,1]$.
To handle the large dynamic range, interpolation is performed in log-space:
\begin{equation}
    \alpha(Q_t) 
    = \alpha_{\min} \left(\frac{\alpha_{\max}}{\alpha_{\min}}\right)^{1-Q_t}
    \label{eq:dr_weight}
\end{equation}
From the linearized system of Eq.~\eqref{eq:ba}, the Hessian takes the form with two parts:
\begin{equation}
    H(Q_t) = J_v^T \Sigma_v^{-1} J_v + \alpha(Q_t) J_d^T \Sigma_{d0}^{-1} J_d\label{eq:adapt_hess}
\end{equation}
The DR term $J_d^T \Sigma_{d0}^{-1} J_d$, introduces constraints on the full 6-DoF pose, 
producing dense block fill-ins in the Hessian. These constraints provide auxiliary support when 
visual measurements are sparse or unreliable, ensuring that the pose graph remains well connected 
and the optimization stable.
The adaptive prior acts as a regularizing term to maintain optimization conditioning when visual tracking fails.

\subsubsection{Scoring System for $Q_t$}\label{sec:tracking_quality}
\begin{figure}[t]
  % \vspace{0.06in}
  \centering
  \begin{tikzpicture}[inner sep=0pt, outer sep=0pt]
    \node[anchor=north] (feat_rmse) at (0,0) 
      {\includegraphics[width=0.5\textwidth]{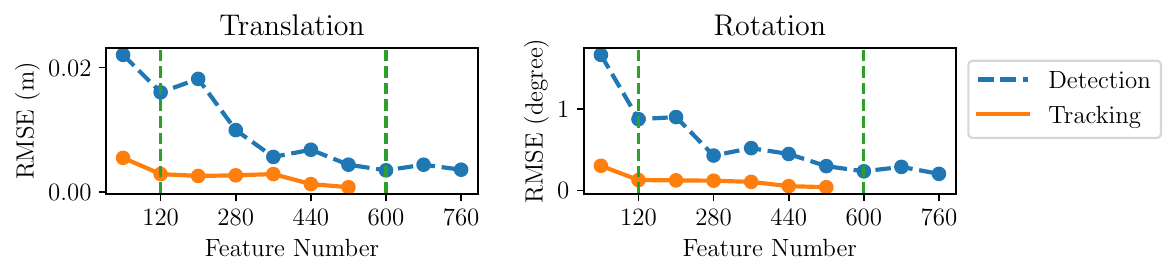}};
    % \node[anchor=west,xshift=0pt,yshift=0pt] (feat_track) at ($(feat_det.east)+(0.0em, 0.0em)$)
      % {\includegraphics[width=0.25\textwidth]{figs/cid_feat_track_rmse.pdf}};
    % \node[anchor=north,xshift=-35pt,yshift=-8pt] (track_score) at ($(feat_det.south)+(3.0em, 0.0em)$)
      % {\includegraphics[width=0.4\textwidth]{figs/cid_apartment_failure.pdf}};
  \end{tikzpicture}
  % Can this be made into a regular line graph(s) instead? (RMSE vs Feature Count, and just scatterplot or connect the dots)
  \caption{Relationship Between Frame Pose RMSE and Tracking Statistics on CID Dataset.
  The detection variable is the number of features detected per frame (at most 800 in the settings).
  The tracking variable is the number of features matched between current frame and local map.
  \label{fig:tracking_quality}}
  \vspace{-10pt}
\end{figure}

Jacobian conditioning can be misleading in visual SLAM, where appearance-based measurements are only indirectly tied to geometry and poor tracking or incorrect data associations can distort measurement quality in low-texture scenarios.
This contrasts with LiDAR SLAM, where depth measurements directly encode geometric structure.

% Relying on Jacobian conditioning as a measure of visual tracking uncertainty can be misleading, since appearance-based tracking does not consistently correlate with the underlying system conditioning. 
% This dependency on feature quality stands in contrast to LiDAR SLAM, where measurements directly capture geometry.
% Relying on Jacobian conditioning as a measure of visual uncertainty can be misleading, since rank-deficient systems may arise even in a \textcolor{blue}{normally operated} SLAM system \cite{Demmel2021SquareRM}. It reflects the strong dependency on feature tracking quality, which is an inherent difference compared to LiDAR SLAM.

Prior work has investigated using the number of tracked features as an indicator \cite{Lee2024SwitchSLAMSL}, providing an early warning of visual tracking failure and predicting degradation before pose estimation collapses.
Consistent with their findings,
evaluation of a vision-only SLAM system \cite{zhao2020good} on the CID dataset \cite{zhang2023cid}, which contains challenging low-texture scenes, 
% correlation exists between the decreased number of detected and tracked features and increased frame-level pose RMSEs.
reveals a negative correlation between feature availability and frame-level pose RMSE (Fig.~\ref{fig:tracking_quality}).
% as the number of detected and tracked features decreases, the frame-level pose RMSE increases .
The effect is more pronounced for the number of detected features, where RMSE grows quadratically once the count drops below 600.
In contrast, the feature tracking count shows lower sensitivity once it exceeds 120, where RMSE continues to decrease but at a diminishing rate.
These results indicate that the detected feature count $N_{det}$ and the tracked feature count $N_{trk}$ serve as reliable indicators of visual tracking quality.

% These observations provide clear evidence that the detected feature number $N_{det}$ and the tracked feature number $N_{trk}$ are reliable anticorrelates with visual tracking RMSE.
To map these tracking variables to a quality score, define
\begin{equation} \label{eq:quality}
    Q_t = \omega_1 \frac{N_{det}}{N_{det}^r} + \omega_2 \frac{N_{trk}}{N_{trk}^r}, 
\end{equation}
% TODO (yanwei): add another figure to show what does the score curve look like.
where $N_{det}^r$ and $N_{trk}^r$ are preset target feature constants for reliable detection and tracking, and $\omega_1$, $\omega_2$ are nonnegative scalars such that $\omega_1 + \omega_2 = 1$.
Each ratio is clipped to $[0, 1]$, ensuring $Q_t \in [0, 1]$.
% The formulation can be naturally extended to include additional metrics, such as conditioning values or measures of feature distribution quality \cite{Ma2025Quali}, by adding additional terms to the score.
Since these two variables are computed during the initial tracking stage, recovering $Q_t$ introduces minimal overhead.
The contribution lies in the proactive incorporation of such visual cues into covariance regulation throughout the SLAM pipeline, enabling more adaptive and reliability-aware state estimation.

\subsubsection{DR Weighting Bounds $\alpha$} \label{sec:dr_weight}
The DR weighting limits $\alpha_{\min}$ and $\alpha_{\max}$ should be such that the DR contribution to optimization is appropriately regulated.
DR should have little to no influence when tracking quality is high ($Q$ large),
and increasingly influential as tracking quality degrades ($Q$ decreases), to dominate the optimization when tracking quality is low ($Q$ small).
Theoretically, finding the bounds can be tied to information-theoretic measures of the pose Hessian, such as \textit{trace}, \textit{log-determinant}, or \textit{minimum eigenvalue} \cite{Carlone2019AttentionAA, zhao2020gfm, zhao2020good}.
These metrics provide a principled way to regulate the balance between DR and visual constraints.
However, such approaches require Hessian computations normally built during optimization, not available before, and their conditioning may not consistently reflect actual system performance.

%% Theretical Analysis, worth trying ???
% Following Eq.~\eqref{eq:adapt_hess}, let
% \begin{equation}
% H_v = J_v^\top \Sigma_v^{-1} J_v, 
% \qquad H_d^0 = J_d^\top \Sigma_{d0}^{-1} J_d,
% \end{equation}
% denote the contributions from visual and nominal DR constraints, respectively.

% we select $\alpha_{\min}$ so that DR contributes only a small fraction of pose information when vision is healthy, and $\alpha_{\max}$ so that DR provides substantial support when vision is weak:
% \begin{align}
% \mathrm{logdet}(\alpha_{\min} H_d^0) 
% \;\approx\; & \rho_{\min}\,\mathrm{logdet}(H_v^{\text{healthy}}) \\
% \mathrm{logdet}(\alpha_{\max} H_d^0) 
% \;\approx\; & \rho_{\max}\,\mathrm{logdet}(H_v^{\text{weak}})
% \end{align}
% with $\rho_{\min} \in [0.05, 0.10]$ and $\rho_{\max} \in [0.4, 0.6]$.

% To ensure well-conditioned optimization, it would also require that the smallest eigenvalue 
% of the Hessian in weak-vision scenarios is lifted above a threshold:
% \begin{equation}
% \lambda_{\min}\!\left(H_v^{\text{weak}} + \alpha_{\max} H_d^0\right) 
% \;\ge\; \tau_\lambda,
% \end{equation}
% where $\tau_\lambda$ is a minimum eigenvalue chosen to prevent degeneracy.

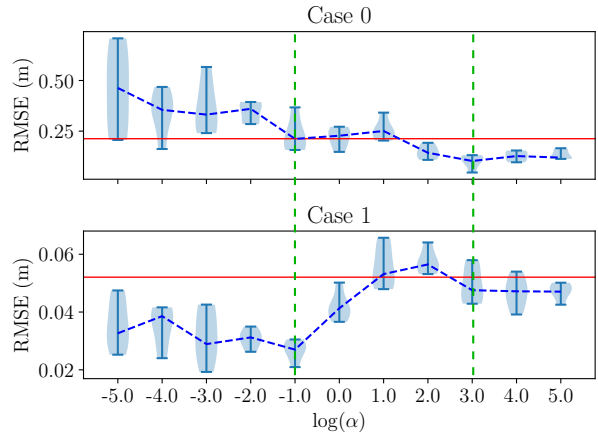
\begin{figure}[t!]
  \begin{tikzpicture}[inner sep=0pt, outer sep=0pt]
  \node[anchor=north west] (dr_weight) at (0,0) 
    {\includegraphics[width=0.45\textwidth]{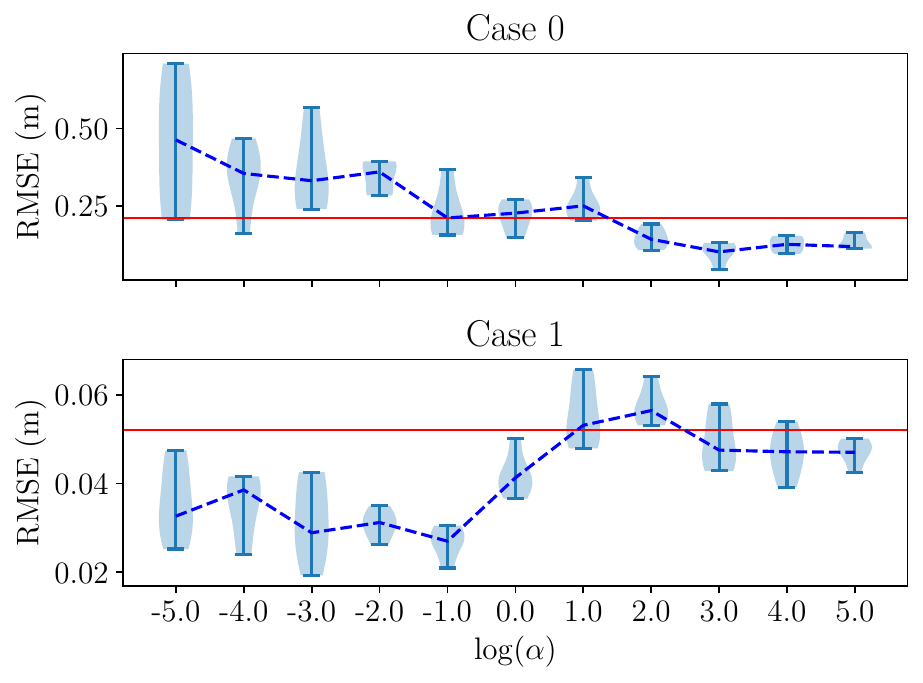}};

  \newcommand{\ftop}{-0.5cm}
  \newcommand{\fbottom}{-4.6cm}
  \coordinate (t1) at ($(dr_weight.north west) + (3.88cm, \ftop)$);
  \coordinate (t2) at ($(dr_weight.north west) + (6.25cm, \ftop)$);
  \coordinate (b1) at ($(t1) + (0cm, \fbottom)$);
  \coordinate (b2) at ($(t2) + (0cm, \fbottom)$);

  \draw[black!30!green, dashed, thick] ($(t1)$) -- ($(b1)$);
  \draw[black!30!green, dashed, thick] ($(t2)$) -- ($(b2)$);
  
  \end{tikzpicture}
  \caption{
  DR weight sweeping study on sequence \textit{floor13\_1}.
  Two segments with poor (case 0) and good (case 1) tracking were evaluated independently, each repeated five times.
  Median RMSE values are connected by the blue dashed line.
  Dotted green lines show the values of $\alpha_{min}$ and $\alpha_{max}$ chosen for DR weighting bounds.
  The solid horizontal red line represents the RMSE achieved by pure dead reckoning (no visual feedback).
  \label{fig:dr_weight}
  % \textcolor{red}{ JCP: If it isn't too hard to generate these sweeps, generating a couple sweeps for other (non calibration) datasets might be good evidence that the empirical method works well}
  }
  \vspace{-10pt}
\end{figure}

% (YD) was the following part clear???
The objective is to achieve robust DR fusion across varying tracking conditions rather than precise weighting.
A lightweight, data-driven strategy is adopted, using the \textit{floor13\_1} sequence of the CID dataset with nominal DR covariances $\Sigma_{d0}$ set to ${0.004}$ \text{m/frame} (translation) and $0.1^{\circ}$\text{/frame} (rotation).
Safe $\alpha$ bounds are obtained through empirical sweeps, with each candidate evaluated by trajectory accuracy (frame-level RMSE) as shown in Fig.~\ref{fig:dr_weight}.
Two segments with poor (case 0) and good (case 1) tracking were evaluated independently, with each test (a single $\alpha$ value) repeated five times.
The results show that in both cases, DR begins to take effect at $\log(\alpha) = -1.0$ and gradually increases its influence, dominating the optimization at $\log(\alpha) = 3.0$.
These values define the chosen operational bounds, 
set to capture the range where DR has a measurable effect on visual estimation rather than globally optimal weights. Avoiding fine-tuning prevents overfitting to specific sequences and preserves generality. Once established, the bounds are fixed across all experiments.
% The bounds are deliberately relaxed, capturing only the shift in DR contribution rather than fine-grained visual quality. Once identified, they are fixed across all experiments, avoiding per-sequence tuning and reducing overfitting.
Together with the adaptive rule $\alpha(Q_t)$ in Eq.~\eqref{eq:dr_weight}, this ensures DR contributes only under visual tracking degradation, while vision remains dominant when reliable.

The next subsection describes how these adaptively weighted DR priors integrate into the different modules of the Visual SLAM pipeline. 
% Specifically, we show their role in the front-end tracking, local bundle adjustment, and loop closing, and explain how the weighting scheme interacts with each component of the system.

\subsection{Integration with Visual SLAM}

\subsubsection{Motion Model for Feature Tracking}\label{sec:da}
As presented in Fig~.\ref{fig:sys_overview} (top block), DR first provides a Motion Model for feature matching. 
Visual pose estimation depends on establishing correct feature correspondences, which are typically obtained by projecting landmarks from the local map into the current frame using a predicted pose. This is fundamentally different from LiDAR-SLAM with geometric association.
In vision-only SLAM systems\cite{murORB2, campos2021orb, zhao2020good}, prediction often employs a constant-velocity (CV) motion model with correspondences limited to a 2D search window centered on predicted projections.
The CV model only extrapolates from past states and does not capture recent motion, making it sluggish and unreliable during rapid maneuvers.
In contrast, DR measures robot kinematic information to generate motion estimates that more closely follow the robot’s short-term trajectory, often at a higher rate than visual estimation, thereby providing stronger motion priors for feature tracking.

\subsubsection{Map-to-Frame Pose Estimation} \label{sec:pe}
\textbf{Track-Local-Map} happens after feature tracking is finished.
A motion-only bundle adjustment (BA) is used for pose estimation, which is a reduced form of the same optimization problem as in Eq.~\eqref{eq:ba}.
Only the current camera pose $T_{t}$ is optimized while landmark positions $X_j$ are fixed:
\begin{equation}
\min_{\mathbf{T}_t}
\sum_{j} \left\| r^{\text{vis}}_{tj}(\mathbf{T}_t, \mathbf{X}_j) \right\|^2_{\Sigma_v^{-1}}
+ 
\left\| r^{\text{DR}}_{t}(\mathbf{T}_{t}, \mathbf{T}_{t-1}) \right\|^2_{\Sigma_d^{-1}}
\label{eq:mba}
\end{equation}
At this stage, robust estimation is typically used to suppress outliers, yet performance still depends on having sufficient inliers.
%In indoor environments, this assumption often fails: plain walls, narrowed navigation space can leave the estimator with few or no valid tracks in previous step.
In indoor environments, this assumption often fails: plain walls and narrow spaces can leave the estimator with few or no detected landmarks.
The problem then becomes under-constrained, and pose estimation quickly diverges.
Since feature detection and tracking statistics are available beforehand, they are used to trigger adaptive DR weighting per Eq.~\ref{eq:dr_cov}, keeping the Hessian well-conditioned and the optimization stable. DR serves as a short-term constraint and motion prior, maintaining state connectivity and ensuring continuous pose output when vision cannot sustain tracking.

\subsubsection{Local Bundle Adjustment} \label{sec:lba}
% A keyframe is created under certain conditions in tracking and passed to local mapping and loop closing for further refinement. Typically, keyframe creation relies on tracking statistics and motion profile, ensuring that new map points are generated and tracked consistently.

% In local mapping, each keyframe is first integrated into the covisibility graph by linking to keyframes that share common map points. It is then included in local bundle adjustment (LBA) for refinement as formulated in Eq.~\eqref{eq:ba}. This process is effective when residuals are sufficiently informative and visual tracking remains reliable.

% In texture-less environments, however, newly created keyframes often exhibit poor covisibility with their neighbors, leading to weak or even missing visual constraints. To mitigate this, we introduce dead reckoning (DR) as additional relative constraints and incorporate temporally adjacent keyframes, thereby reinforcing connectivity and ensuring that the LBA problem remains well conditioned, as presented in Fig.~\ref{fig:dr_ba}.

A keyframe is created based on tracking statistics and motion profiles, then refined through local bundle adjustment (LBA) by linking to its covisible neighbors. While this process works under reliable visual tracking, texture-less environments often yield poor covisibility and weak or missing constraints. To address this, we incorporate DR as additional relative constraints to temporally adjacent keyframes, thereby reinforcing graph connectivity and keeping the LBA problem well conditioned.

\begin{figure}[t]
  % \vspace{0.06in}
  \centering
  \includegraphics[width=1.0\linewidth]{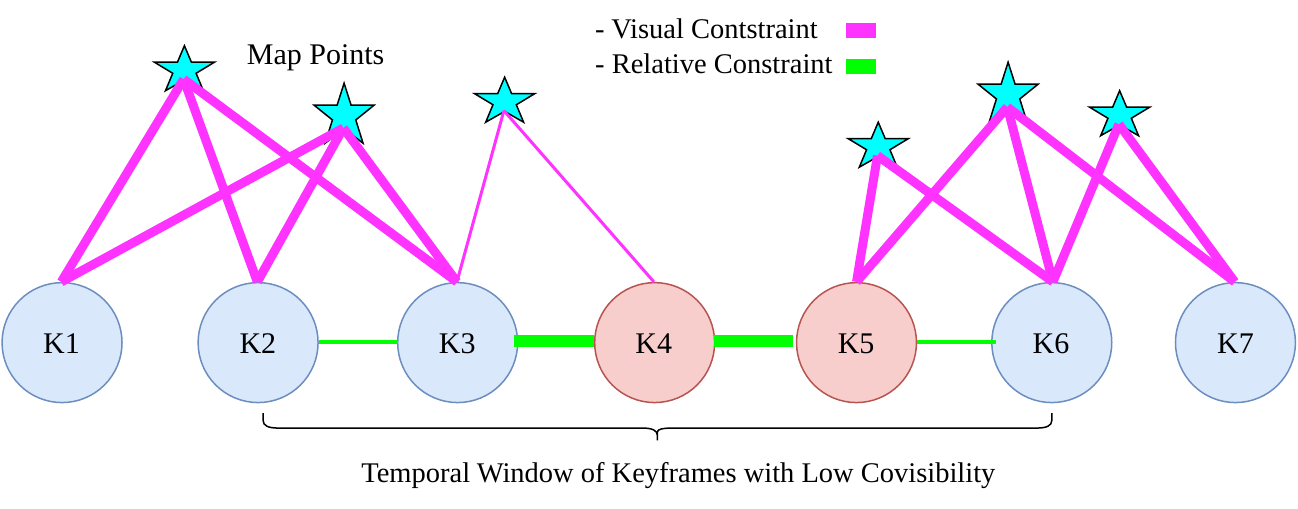}
  \caption{Adaptive DR weighting in bundle adjustment. Stronger DR constraints (wider lines) are applied to low-covisibility keyframes ($K4$, $K5$) and their neighbors, then fade out as tracking recovers.
  \label{fig:dr_ba}}
  % \vspace{-15pt}
\end{figure}

The weight of DR constraints at this stage is scaled according to the number of covisibility connections. 
Let $C_{ij}$ denote the connection count (number of visual edges) between keyframes $i$ and $j$.
$C_{ref}$ is defined as a reference connection value, estimated online during local bundle adjustment (LBA). Specifically, $C_{ref}$ is computed as the median connection count from well-tracked keyframes in the most recent covisibility graph, 
representing the typical connection strength under reliable tracking conditions.
The keyframe quality score for LBA is then defined as,
\begin{equation}
    Q_{ij} = \frac{C_{ij}}{C_{ref}}
\end{equation}
The ratio is saturated so that $Q_{ij}$ remains bounded within $[0, 1]$.
An initial weight value is estimated as per Eq.~\ref{eq:dr_weight}, such that poorly connected keyframes receive stronger DR influence.
Applying a windowed smoothing strategy distributes DR weights to adjacent keyframes. The weighting decreases with covisibility distance, giving closer neighbors higher weights and ensuring a gradual, consistent influence of DR across the local window, as illustrated in Fig.~\ref{fig:dr_ba}.

\subsubsection{Global Bundle Adjustment} \label{sec:ba}
Global bundle adjustment (GBA) is triggered upon loop closure and jointly refines all keyframe poses and landmarks to correct long-term drift and achieve a globally consistent map.
The same connectivity issues observed in LBA also appear at the global level, where low-covisibility keyframes contribute little to the optimization.
To address this, DR-stabilized weights from LBA are preserved, keeping the global optimization in Eq.~\ref{eq:ba} well posed.
This stabilization is particularly important for robotic applications like inspection and surveying \cite{Staniaszek2025AutoInspect, lattanzi2017review}, where repeat traversals are frequent.
In such scenarios, a corrected and consistent map allows frame-level poses to align with keyframe-level accuracy, providing reliable localization for robust long-term navigation.

Employing DR throughout the hierarchical design preserves system conditioning for all optimizations, which delivers continuous and accurate pose estimation even in challenging scenarios.
This is most effective during short-term visual tracking losses, where DR maintains connectivity and prevents failure. For extended visual outages, the accumulated error is limited by DR accuracy.

%% file: exp.tex
Validation of the proposed method involved integrating the adaptive weighting modules into a feature-based visual SLAM system \cite{zhao2020good}.
% , consisting of a low-latency front-end (Good-Feature) and a budget-aware back-end (Good-Graph).
% \textcolor{red}{JCP: No point to mentioning these features. We aren't here to discuss GF-GG. To reduce verbosity, maybe shorten it to just GG (the paper cited is the good-graph paper).}
Dead reckoning (DR) is implemented using an EKF-based framework following \cite{lynen2013robust}, fusing in robot odometry information.
The system supports stereo and RGB-D modes, where for map point initialization the former uses stereo matching and the latter uses the depth measurement.
% Unlike some RGB-D SLAM systems \cite{RGBD-SLAM}, the depth measurements are not further exploited during subsequent processing.

Denote the full Good Weights method as {\gw} (DR in all modules) and {\gw}:\textit{X} to indicate variants with DR applied incrementally to each module.
\textit{DA} refers to feature data association (\S\ref{sec:da}), \textit{PE} to pose estimation in tracking (\S\ref{sec:pe}), 
and LBA to local bundle adjustment (\S\ref{sec:lba}).
{\gw} parameters are fixed for all evaluations as follows: 
$w_1 = 0.5$, $w_2 = 0.5$, $N_{det}^r = 600$, and $N_{trk}^r = 120$ (\S\ref{sec:tracking_quality}, Fig.~\ref{fig:tracking_quality}); 
$\alpha_{min} = 10^{-1}$ and $\alpha_{max} = 10^{3}$ (\S\ref{sec:dr_weight}, Fig.~\ref{fig:dr_weight}); 
and $C_{ref} = 20$ for initialization then adjusted online during local bundle adjustment (\S\ref{sec:lba}).
To avoid overfitting, parameter tuning involved only sequence \textit{floor13\_1}, with the resulting parameter settings kept fixed across all
other sequences, datasets, and robot experiments.
% prioritizing robustness under visual degradation over dataset-specific optimization.

% The parameters used in our experiments are set as follows and fixed all the testings: 
% $w_1 = 0.5$, $w_2 = 0.5$, $N_{det}^r = 800$, $N_{trk}^r = 300$ in Sec.~\ref{fig:tracking_quality}, Fig.~\ref{fig:tracking_quality}, 
% $\alpha_{min} = 10^{-3}$, $\alpha_{max} = 10^{4}$ as in Sec.~\ref{sec:dr_weight}. 
% and $C_{ref} = 20$  as initial value and adjusted online as in Sec.~\ref{sec:ba}.
% These values were determined empirically from the analysis in Sec.~\ref{sec:meth} (see Fig.~YY), where parameter sweeps revealed stable operating ranges. Once identified, the values were fixed for all experiments to ensure consistency and to avoid sequence-specific tuning.

% \textcolor{red}{JCP: All parameters that haven't been defined yet ($\gamma, w_1, w_2, N_{det}^r, N_{trk}^r,C_{ref}$) should be defined (preferably with justification for why they were chosen, or explicitly stating that it was an arbitrary choice.) If not here, then in an appendix.}

\subsection{CID Open-Loop Benchmarking}
The evaluation is first conducted on the public CID benchmark sequences \cite{zhang2023cid}, an indoor dataset with synchronized RGB-D images, IMU, and wheel odometry measurements, with Ground-truth generated using a high-accuracy LiDAR SLAM system. It contains 22 challenging sequences with diverse motion profiles and low-texture scenes.
Two multi-floor sequences were excluded given that the data capture device was lifted to change floors and the method currently assumes available wheel odometry.
The {RGBD} baseline methods are: {\orb} \cite{campos2021orb} in RGBD–Inertial mode, {\groundfusion} \cite{yin2024ground} as an RGBD-Inertial–Wheel odometry system, 
and {\pgd} \cite{zhang2025pgd}, a plane-enhanced RGBD-Inertial system, together with its variants: \textit{w/o P} with only point features, and \textit{w/o G} with point and planar features.
Since the {\pgd} code is not public, reported performance is from \cite{zhang2025pgd}.

\begin{figure}[t]
  % \vspace{0.06in}
  \centering
  \begin{tikzpicture}[inner sep=0pt, outer sep=0pt]
    \node[anchor=north] (feat_det) at (0,0) 
      {\includegraphics[width=0.45\textwidth]{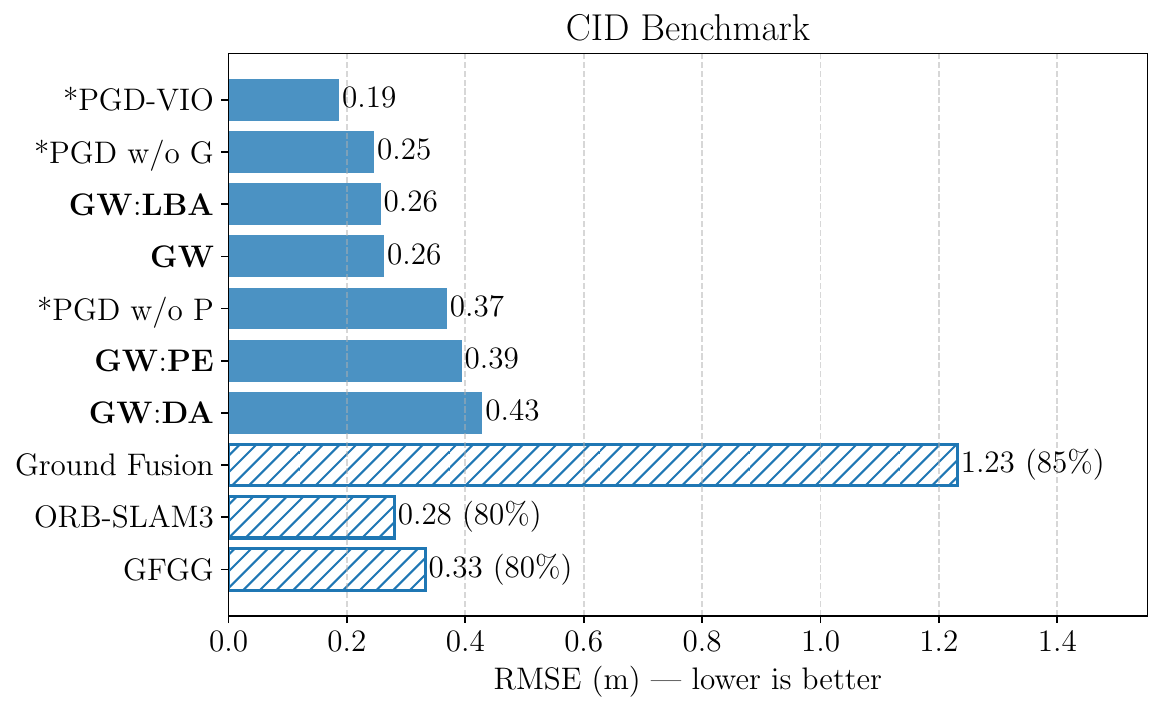}};
    % \node[anchor=west,xshift=0pt,yshift=0pt] (feat_track) at ($(feat_det.east)+(0.0em, 0.0em)$)
      % {\includegraphics[width=0.25\textwidth]{figs/cid_feat_track_rmse.pdf}};
  \end{tikzpicture}
  \caption{
  Performance comparison of SLAM methods on the CID dataset. 
  Methods are sorted from top to bottom by sequence completeness and then by RMSE. 
  Results marked with * are reported from \cite{zhang2025pgd}, 
  as the implementation is not publicly available.
  The proposed methods are shown in bold. 
  Hashed lines indicate incomplete sequence results, with the value given in parentheses next to the RMSE.
  % \textcolor{red}{JCP: What do the hashed lines mean? Your methods should be bolded. Based on the ablation study, is GFGG-DR only an improvement over GFGG-DR-LBA in repeat runs? That's odd...}
  % \textcolor{red}{More importantly, GFGG-Base is an important ablation here.}
  \label{fig:cid_bench}}
\end{figure}

Evaluation follows the protocol in \cite{zhang2025pgd}:
Tracking accuracy is evaluated using APE RMSE over five runs per sequence. 
A sequence is marked as failed if any run has RMSE above 10 m or a tracking ratio below 50\%.
Fig.~\ref{fig:cid_bench} depicts the sorted completeness (ratio of completed sequences to total) with secondary sorting by average RMSE.
All experiments are performed on an Intel i7-12700K processor.
% (single-core passMark score: 4011).

\subsubsection{Result Analysis}

\paragraph{Frame Tracking Performance}
From bottom to top, 
{\gfgg}, {\orb} and {\groundfusion} fail to achieve full completeness (80\%, 80\% and 85\%, respectively), 
suggesting that vision-only systems and tightly coupled sensor-fusion approaches lack robustness under these conditions.
In low-texture environments, poor visual tracking forces inertial or wheel-odometry cues 
to dominate the optimization; however without sufficient visual correction they accumulate drift, leading to degraded accuracy.
For {\gw}, full completeness is first achieved when DR is applied to feature data association (DA), 
with further accuracy gained when extended to pose estimation (PE). 
In both cases, DR stabilizes the front-end tracking thread, strengthening feature association and ensuring continuous and robust pose estimation.
When integrated into the back-end, both {\gw}:LBA and {\gw} achieve an RMSE of $0.26$ m, representing a 35\% improvement over the front-end variants.

The performance of {\gw} closely matches the planar variant of {\pgd} \textit{w/o G} (RMSE: 0.25 m) and outperforms the point-only baseline (\textit{w/o P}) by 29\% (RMSE: 0.37 m).
The full {\pgd} with graph-based drift suppression achieves a lower error of 0.19 m, reflecting its explicit geometric registration and drift correction in highly planar indoor environments.
In contrast, {\gw} relies only on point reprojection error, reinforced with adaptively weighted DR according to tracking quality, yet still reaches accuracy on-par with plane-based methods without the added complexity of plane extraction or graph maintenance for plane matching.
This shows that a lightweight, adaptively DR-augmented point-based visual SLAM can deliver competitive accuracy while remaining practical for real-robot deployment.
% \textcolor{red}{
\paragraph{Runtime}
% TODO: If space allowed, add a table.
Frame tracking latency measure the time from image reception to pose output. {\gw} achieves an average latency of 14.84 ms, compared to 15.52 ms for the baseline {\gfgg}, indicating no additional latency overhead while providing improved robustness and accuracy in challenging scenarios.
% }

\paragraph{Global-BA and Map-based SLAM Performance}
To further assess the effect of global bundle adjustment with DR (\S\ref{sec:ba}) and to present how the visual map contributes to SLAM accuracy, the repeat-run protocol from \cite{du2024task} is adopted.
In this setup, each sequence is executed for three continuous loops without resetting the system, 
so the visual map constructed is retained and reused for the subsequent runs, and the frame-level RMSE is then computed over these additional loops.
Table.~\ref{tab:cid-repeat-run} summarizes the average RMSE across CID sequences, along with the ratio between frame and keyframe RMSE, defined as $R(F/KF) = \frac{Frame\ RMSE}{KeyFrame\ RMSE}$, 
This ratio reflects how closely frame poses align with the optimized keyframe poses maintained in the SLAM map; values closer to 1.0 indicate that frame accuracy is nearly as good as keyframe accuracy, demonstrating stronger map consistency.

\begin{table}[t!]
    \centering
    \caption{Repeat-Run Evaluation on CID. 
    % Results better than {\pgd} (0.19 m) are highlighted in bold.
    }
    \label{tab:cid-repeat-run}
    \begin{tabular}{|>{\bfseries}c | c c || c c |}
    \hline
    \multirow{2}{*}{Method} &
    \multicolumn{2}{c || }{Frame RMSE (m)} &
    \multicolumn{2}{c | }{ R(F/KF)} \\
                  & Loop-2 & Loop-3 & Loop-2  & Loop-3 \\ \hline
    {\gw}:DA  & 0.274  & 0.333  & 2.920   & 4.312  \\ \hline
    {\gw}:PE  & 0.246  & 0.234  & 1.092   & 1.032  \\ \hline
    {\gw}:LBA & 0.243  & 0.242  & 1.092   & 1.074  \\ \hline
    {\gw}     & \textbf{0.176} & \textbf{0.169} & 1.092 & 1.053 \\ \hline
    \end{tabular}
    % \begin{tabular}{|>{\bfseries}c | c c c c|}
    % \hline
    % Method  & {\gw}:DA & {\gw}:PE & {\gw}:LBA & {\gw}  \\ \hline
    % Loop-1  &  0.246  & 0.176	 & 0.16   & 0.0  \\ \hline
    % Loop-2  &  0.246  & 0.176	 & 0.169  & 0.0  \\ \hline
    % Loop-3  &  0.246  & 0.176	 & 0.169  & 0.0  \\ \hline
    % \end{tabular}
\end{table}

The first observation is that {\gw} achieves RMSEs of $0.176$ m and $0.169$ m in the second and third loops, respectively.
Frame-level tracking errors converge toward keyframe accuracy, with the error ratio approaching 1.0.
This demonstrates the benefit of reusing the SLAM map in passive visual SLAM, allowing frame poses to be refined to near keyframe-level accuracy.
% Such consistency is particularly valuable for robotic applications like inspection and surveying, where repeated traversals are common \cite{lattanzi2017review, Staniaszek2025AutoInspect}.
DR in {\gw}:PE and {\gw}:LBA shows a similar trend but with reduced accuracy compared to {\gw}, largely because weak visual constraints yield poor keyframe connectivity and a less stable BA problem.
The {\gw}:DA variant, which applies DR only at the front-end for feature tracking, provides little benefit in low-texture scenarios with no visual measurements and can lead to incorrect estimates.
These observations highlight the importance of integrating DR information consistently across the hierarchical modules of visual SLAM to keep the system well-conditioned.  Of note, repeat-run, map reuse {\gw} performance matches that of first run PGD-VIO (recall that PGD-VIO explicitly uses depth information).
Overall, the results show that with DR assistance, visual SLAM can reliably complete trajectory tracking on the first pass and then exploit the constructed map by adaptively balancing DR and visual constraints, which is a capacity that conventional odometry-only fusion systems generally lack.

\subsection{Real Robot Navigation Sequences} \label{exp:tsrb}

\begin{figure}[t!]
  % \vspace{0.06in}
  \centering
  \begin{tikzpicture}[inner sep=0pt, outer sep=0pt]
  \node[anchor=north] (seq) at (0, 0)
    {\includegraphics[width=0.45\textwidth]{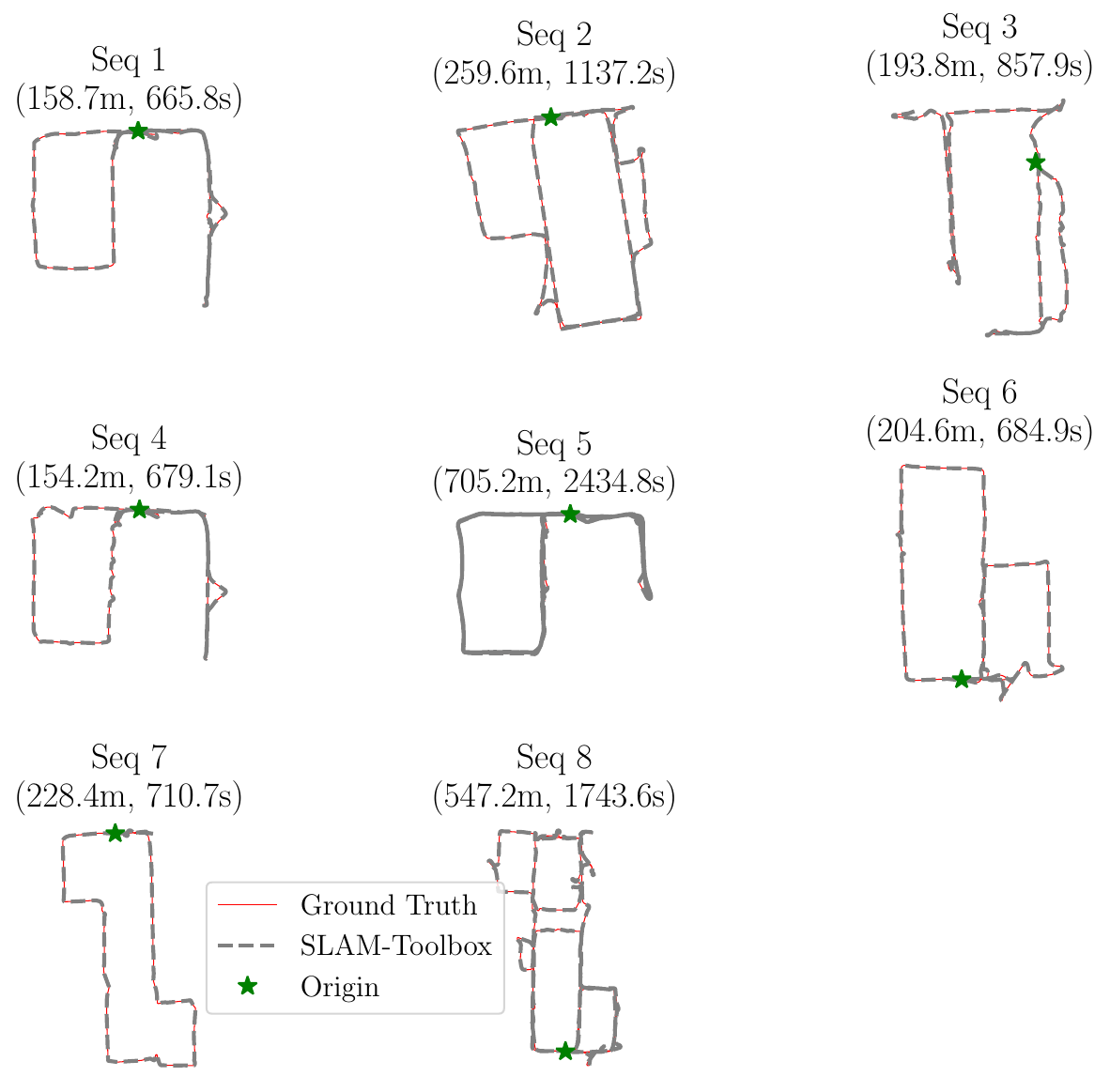}};
  \node[anchor=north, rotate=-90] (robot) at ($(seq.south east)+(-1.5em, 3.0em)$)
    {\includegraphics[width=0.12\textwidth]{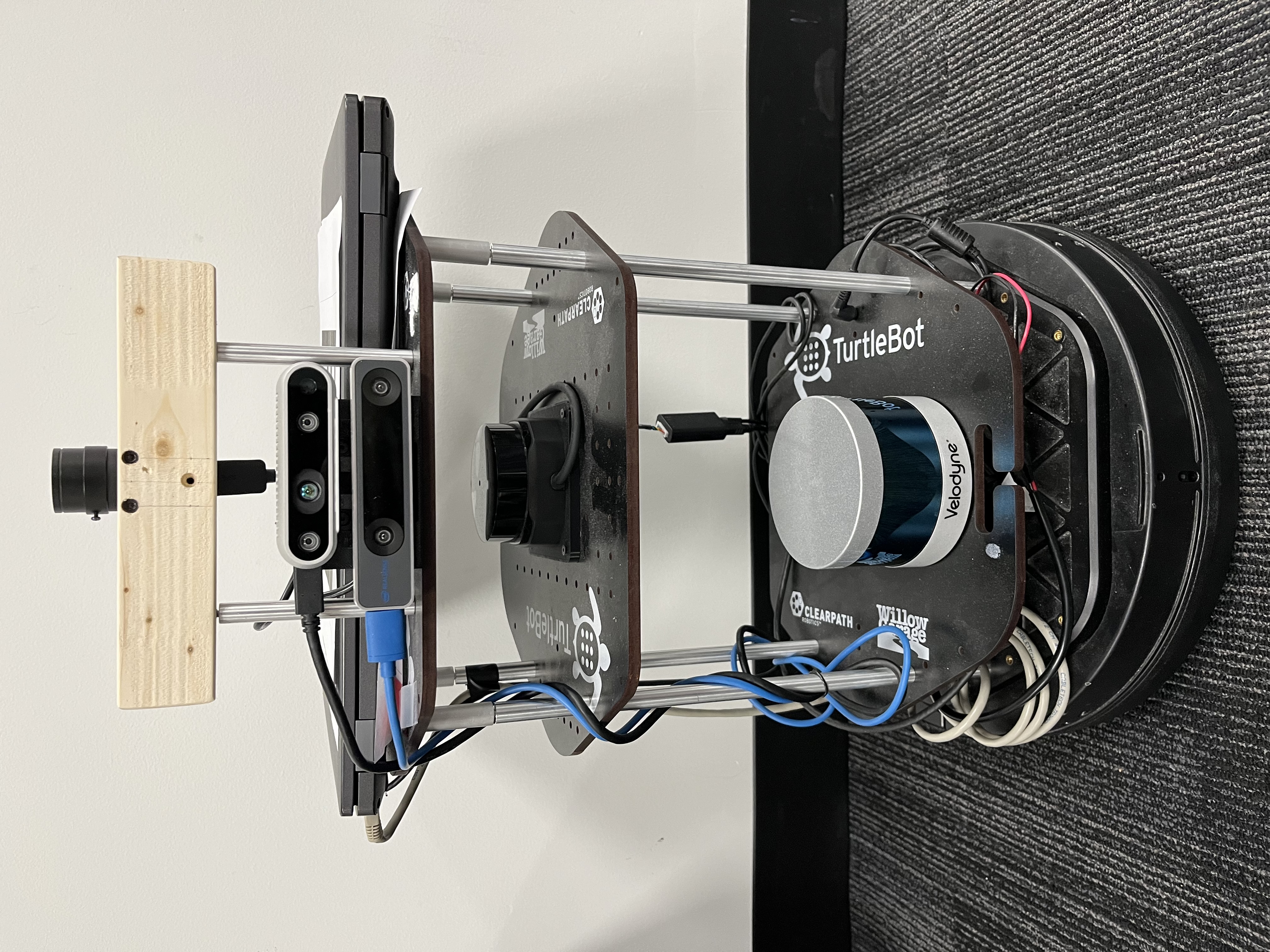}};
  \end{tikzpicture}
  \caption{
  Trajectories of sequences collected in an office environment, with the robotic platform shown in the bottom-right corner.\label{fig:tsrb_seqs}\label{fig:turtlebot}}
  % \vspace{-15pt}
\end{figure}
% We further extend the evaluation to self-collected indoor sequences on our robot platform, demonstrating the real-world applicability of the proposed method for navigation tasks.
The navigation sequences were collected using a TurtleBot2 equipped with a RealSense D435i (stereo and depth at 30 Hz, IMU at 400 Hz) and onboard robot odometry at 20 Hz from a 3-axis gyroscope and wheel encoders. Ground-truth trajectories were obtained with a 2D LiDAR (RPLIDAR-S2, 10 Hz) using SLAM-Toolbox \cite{Macenski2021} and refined through spline-based smoothing \cite{Mueggler2017ContinuousTimeVO}.

In contrast to the CID dataset, the navigation sequences were gathered under fully autonomous navigation with {\slamtoolbox} and the complete ROS navigation stack \cite{ROSNavStack}, following the methodology in \cite{du2024task}. Waypoints were either transmitted remotely or predefined on a floor map.
In total, eight sequences were recorded, covering both single-round exploration and multiple re-visits, as illustrated in Fig.~\ref{fig:tsrb_seqs}.

\begin{figure}[t]
  % \vspace{0.06in}
  \centering
  \begin{tikzpicture}[inner sep=0pt, outer sep=0pt]
    \node[anchor=north] (feat_det) at (0,0) 
      {\includegraphics[width=0.4\textwidth]{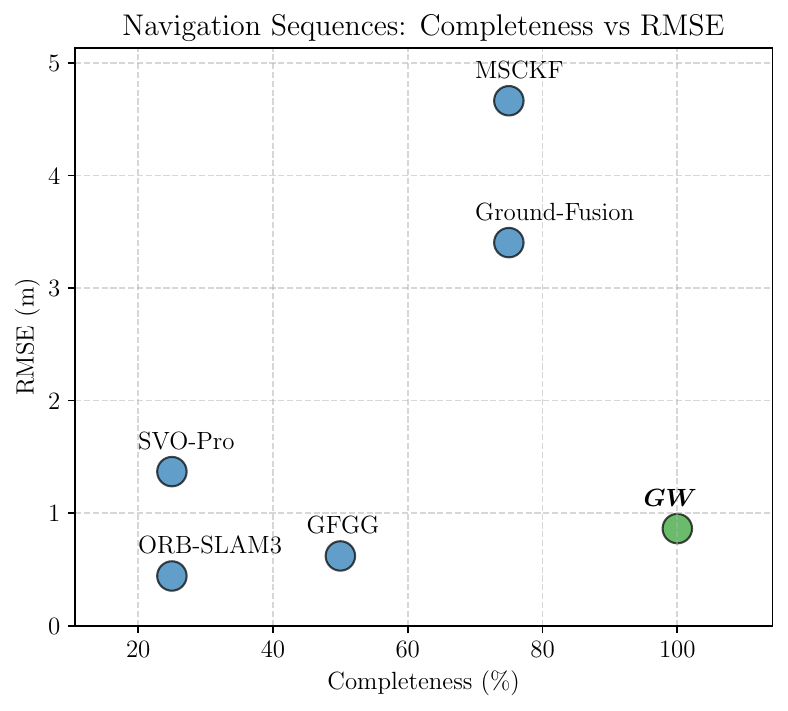}};
    % \node[anchor=west,xshift=0pt,yshift=0pt] (feat_track) at ($(feat_det.east)+(0.0em, 0.0em)$)
      % {\includegraphics[width=0.25\textwidth]{figs/cid_feat_track_rmse.pdf}};
  \end{tikzpicture}
  \caption{Performance comparison of SLAM methods on navigation sequences.
  \label{fig:tsrb_bench}}
  \vspace{-15pt}
\end{figure}

% where passive visual SLAM typically fails due to lack of reliable features. 
Baseline methods include additional established Stereo Visual(-Inertial) SLAM/Odometry systems: {\orb} \cite{campos2021orb}, {\gfgg} \cite{zhao2020good}, \msckf \cite{mourikis2007multi}, {\svo} \cite{qu2022dsol}, \dsol \cite{qu2022dsol} and {\groundfusion} (RGBD-Inertial-Wheel) \cite{yin2024ground}.
% and \textcolor{red}{{\macvo} \cite{Qiu2024MACVOMC}}.
Performance measures are sequence average RMSE and Completeness in Fig.~\ref{fig:tsrb_bench}. PGD-VIO is not available for benchmarking.
% \textcolor{red}{MAC-VO is  missing!}

\subsubsection{Result Analysis}
{\gw} is the only method to achieve full trajectory completeness, with an RMSE of $0.87$ m. 
Compared to the vision-only baseline {\gfgg}, completeness more than doubles while maintaining similar accuracy, 
confirming the effectiveness of incorporating DR.
Focusing on completeness, the next best results are {\msckf} and {\groundfusion} at 75\%, with RMSE values of $4.6$ m and $3.4$ m, respectively.
These results strengthen the benefits of fusing inertial and wheel odometry with visual odometry, 
yet they still fall short of {\gw} due to the lack of adaptive weighting and insufficient exploitation of visual maps when reliable.
{\svo} and {\orb} perform the worst, failing frequently in long or low-texture sequences, demonstrating limited robustness for real robot navigation.
{\dsol} consistently fails across all sequences, likely due to drift accumulation inherent in direct odometry methods; it is excluded from comparison.

% \subsection{Runtime Analysis}

\begin{figure}[t!]
  % \vspace{0.06in}
  \centering
  \begin{tikzpicture}[inner sep=0pt, outer sep=0pt]
    \node[anchor=north] (traj) at (0, 0)
      {\includegraphics[width=0.48\textwidth]{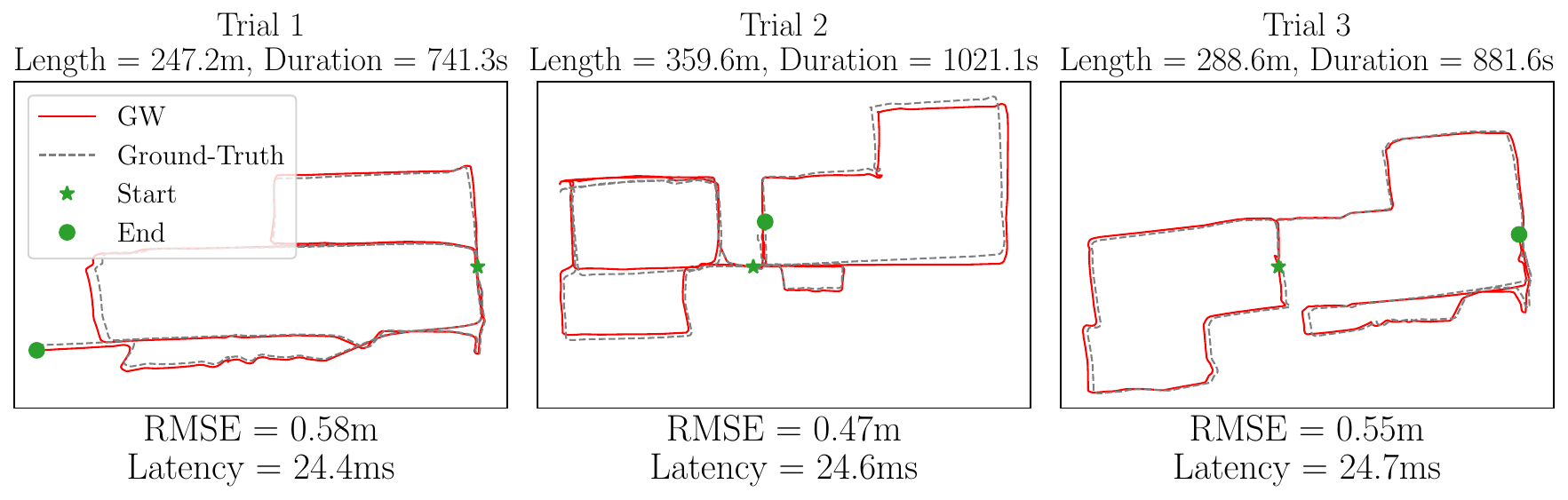}};
    \node[anchor=north] (t1) at ($(traj.south)+(-8.5em, -0.2em)$)
      {\includegraphics[width=0.15\textwidth]{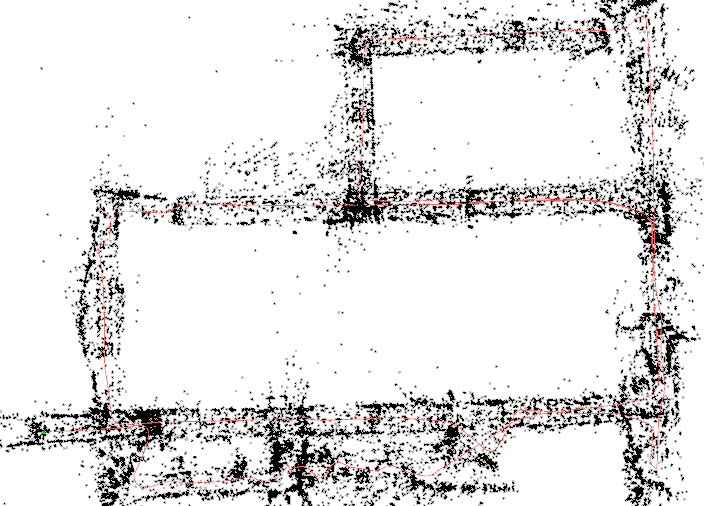}};
    \node[anchor=north] (t2) at ($(traj.south)+(0em, -0.2em)$)
      {\includegraphics[width=0.15\textwidth]{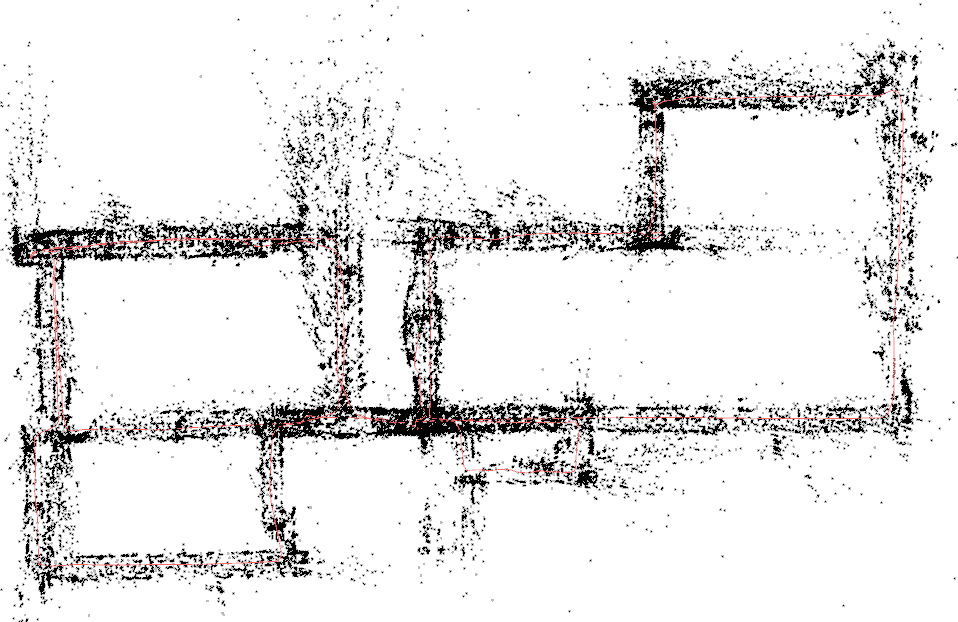}};
    \node[anchor=north] (t3) at ($(traj.south)+(8.5em, -0.2em)$)
      {\includegraphics[width=0.15\textwidth]{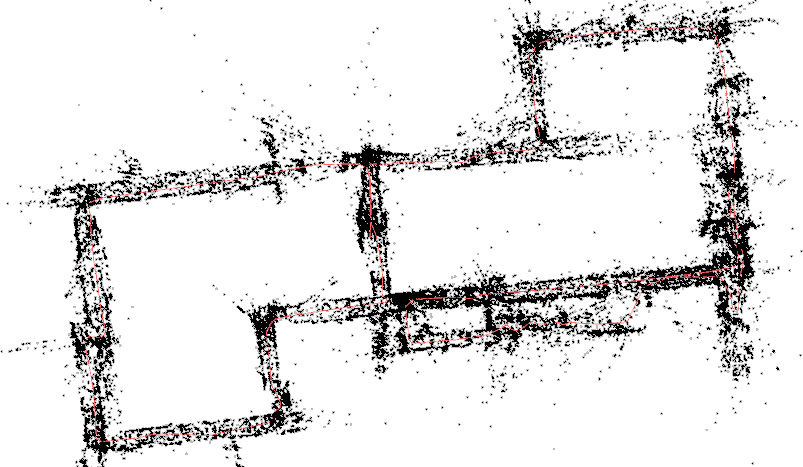}};
  \end{tikzpicture}
  \caption{Closed-loop navigation evaluation in an office environment. 
  Top row: estimated trajectories (red solid) against ground truth (gray dashed) for three distinct navigation trials, with trajectory RMSE (m) and median tracking latency (ms) below. 
  Bottom row: visual maps reconstructed during the corresponding trials.\label{fig:tsrb_cl}}
  % \vspace{-15pt}
\end{figure}

\subsection{Real-Robot Closed-loop Navigation Performance}
To validate that the open-loop outcomes translate to actual deployment, {\gw} is tested on real robot closed-loop navigation tasks 
with all modules running fully onboard a MiniPC of AMD Ryzen 7 8845HS processor.
Using the navigation stack described in Sec.~\ref{exp:tsrb}, the SLAM component is replaced with the proposed method. 
LiDAR scans are recorded during operation and later used to generate ground-truth trajectories offline with {\slamtoolbox}.
Baseline methods are not included, as they failed to complete open-loop trajectories. 
% We report RMSE across closed-loop trials in Fig.~\ref{fig:tsrb_cl}, which demonstrates that our method maintains accuracy and continuity sufficient for reliable navigation.
Testing involved three navigation trials with different navigation paths in an office environment.
Fig.~\ref{fig:tsrb_cl} shows the trajectories (top row) and the corresponding reconstructed maps (bottom row). 
Across all trials, GW achieved stable navigation performance, with an average trajectory RMSE of 0.54 m and a tracking latency of 24.6 ms, while operating in real time on 30 Hz image streams.
The consistent visual maps and low trajectory errors confirm that the proposed approach is reliable for practical navigation tasks.

%% Task-Driven Bench
% Following the methodology in \cite{du2024task}, we use a precision metric that measures the consistency of the robot in repeatedly reaching the same waypoints.
% The robot poses at each waypoint was captured through an upward-facing camera mounted on the robot observing AprilTag attached on the ceiling on top of each waypoihnts.
% Unlike \cite{du2024task}, where waypoints were predefined on a floor map, here waypoints are extracted directly from the pose graph of the SLAM map built from the previous open-loop navigation sequence $S7$.
% Waypoint coordinates may differ between methods, but the precision metric is defined independently of the global coordinate system and is consistent within each map. 
% Specifically, a waypoint is defined when a keyframe is associated with an AprilTag observed by the upward-facing camera during traversal.
% The same set of tags is used by both methods to ensure consistency, and precision is evaluated in a repeat-run manner.
% Particularly, the pipline is as in Fig.~\ref{fig:cl_pipeline}

%% file: conclusion.tex
This paper describes  \textit{Good Weights}, a visual SLAM framework for proactive, adaptive integration of dead-reckoning priors into the tracking, local mapping, and global bundle adjustment modules. 
Unlike LiDAR-based systems, which benefit from dense geometric constraints and rely on post-optimization corrections such as reweighting or smoothing, visual SLAM depends on sparse feature associations and becomes ill-conditioned when those associations fail. 
\textit{Good Weights} addresses this structural limitation by regulating the influence of dead reckoning (DR) based on visual tracking quality to ensure well-conditioned optimization throughout the pipeline. Experiments on public benchmarks and real robot datasets show that this design 
surpasses existing DR-aided visual SLAM approaches and achieves repeat-run performance comparable to depth-based SLAM. 
%In future, the framework could be extended to multi-modal settings, where adaptive priors are managed jointly across vision, inertial, and depth sensors to further enhance robustness in visually and structurally diverse settings.
% \textcolor{red}{FIXME:
% short-term failures.
% }